\documentclass[a4paper,11pt]{article}
\usepackage[utf8]{inputenc}
\usepackage{amsmath,geometry,amssymb,booktabs,tablefootnote}
\usepackage{tikz,pgfplots,multirow}
\pgfplotsset{compat=newest}
\definecolor{codecolor}{HTML}{963611}
\newcommand{\code}[1]{{\raisebox{0pt}{\ttfamily\color{codecolor}#1}}}
\usepackage[breaklinks=true,colorlinks,citecolor=black,urlcolor=black,linkcolor=black]{hyperref}
\newcommand{\forceindent}[1]{\leavevmode{\parindent=#1em\indent}}

\title{MLPGradientFlow:\\going with the flow of multilayer perceptrons\\ (and finding minima fast and accurately)}
\author{Johanni Brea, Flavio Martinelli, Berfin Şimşek, Wulfram Gerstner}

\begin{document}

\maketitle

\begin{abstract}
    MLPGradientFlow is a software package to solve numerically the gradient flow differential equation $\dot \theta = -\nabla \mathcal L(\theta; \mathcal D)$, where $\theta$ are the parameters of a multi-layer perceptron, $\mathcal D$ is some data set, and $\nabla \mathcal L$ is the gradient of a loss function.
    We show numerically that adaptive first- or higher-order integration methods based on Runge-Kutta schemes have better accuracy and convergence speed than gradient descent with the Adam optimizer.
    However, we find Newton's method and approximations like BFGS preferable to find fixed points (local and global minima of $\mathcal L$) efficiently and accurately.
    For small networks and data sets, gradients are usually computed faster than in pytorch and Hessian are computed at least $5\times$ faster.
    Additionally, the package features an integrator for a teacher-student setup with bias-free, two-layer networks trained with standard Gaussian input in the limit of infinite data.
    The code is accessible at \url{https://github.com/jbrea/MLPGradientFlow.jl}.
\end{abstract}

We consider multi-layer perceptrons
\begin{align}
    f_\theta: \mathbb R^{D_\mathrm{in}}&\to\mathbb R^{D_\mathrm{out}}\\
    x&\mapsto f(x) = f^L_{\theta^L} \circ f^{L-1}_{\theta^{L-1}} \circ \cdots  \circ f^1_{\theta^1}(x)\\
    f^i_{\theta^i = (W^i, b^i)}: \mathbb R^{D_{i-1}}&\to\mathbb R^{D_i}\\
    x&\mapsto \left\{\begin{array}{ll}\sigma^i(W^i x + b^i) & \mbox{if } \chi_i = \mbox{true}\\
                                      \sigma^i(W^i x) & \mbox{otherwise}
                                  \end{array}\right.\, ,
\end{align}
with weights $W^i\in \mathbb R^{D_i\times D_{i-1}}$, bias indicator $\chi^i\in\{\mbox{true, false}\}$, biases $b^i\in \mathbb R^{D_i}$ and element-wise applied activation function $\sigma^i:\mathbb R \to\mathbb R$, $D_0 = D_\mathrm{in}, D_L = D_\mathrm{out}$ and $\theta = (\theta^1, \ldots, \theta^L) = (W^1, b^1, \ldots, W^L, b^L)$.
With data $\mathcal D = ((x_1, y_1), (x_2, y_2), \ldots, (x_N, y_N))$ and single-sample loss $\ell(f_\theta(x_i), y_i) \in \mathbb R$, we define the loss
\begin{equation}
    \mathcal L(\theta; \mathcal D) = \sum_i^N\ell(f_\theta(x_i), y_i) + R(\theta)
\end{equation}
with barrier function
\begin{equation}
    R(\theta) = \left\{\begin{array}{cl} 0 & \mbox{if } \frac12\|\theta\|_2^2 \leq c\\\big(\frac12\|\theta\|_2^2 - c\big)^2 & \mbox{otherwise}
\end{array}\right.
\end{equation}
where $c\in\mathbb R_+$ is some constant.
The barrier function is included to guarantee fixed points with finite norm of the parameter values, unless $c$ is chosen to be infinite.
We define the gradient flow differential equation
\begin{equation}\label{eq:gradientflow}
    \dot \theta(t) = -\eta\nabla \mathcal L(\theta(t); \mathcal D)\, ,
\end{equation}
where $\eta$ is a constant and $\nabla\mathcal L$ the gradient of the loss in Euclidean metric.
We aim at finding solutions
\begin{equation}
    \theta(T) = \int_0^T\dot \theta(t)dt
\end{equation}
of the gradient flow differential equation (\autoref{eq:gradientflow}) for initial conditions $\theta(0) = \theta_0$.

If the target values are generated by a multi-layer perceptron, i.e. $y_i = f_{\theta^*}(x_i)$ with parameters $\theta^*$, we speak of a (noise-free) teacher-student setup.
Note that the teacher does not need to have the same number of parameters as the student.

\section*{Results}
MLPGradientFlow (\url{https://github.com/jbrea/MLPGradientFlow.jl}) is written in Julia, but it can also be used and was tested with a Python interface.
The most important functions are listed in \autoref{tab:api}.

\begin{table}[t]
    \footnotesize\centering
    \begin{tabular}{p{.25\textwidth}p{.6\textwidth}}
    finite data constructor\newline & \code{Net(layers = $((D_1, \sigma^1, \chi_1), \ldots, (D_L, \sigma^L, \chi_L))$,}\newline\code{\forceindent{1.5} input = $(x_1, \ldots, x_N)$, target = $(y_1, \ldots, y_N)$)}\\
        infinite data constructor\newline & \code{NetI($\theta_\mathrm{student}$, $\theta_\mathrm{teacher}$)}\\\midrule
        loss and gradient  & \code{loss(net, $\theta$)}, \code{gradient(net, $\theta$)}\\
        Hessian and its spectrum & \code{hessian(net, $\theta$)}, \code{hessian\_spectrum(net, $\theta$)}\\
        solving \autoref{eq:gradientflow} & \code{train(net, $\theta$)} \end{tabular}
    \caption{\textbf{Most important functions of MLPGradientFlow.}}\label{tab:api}
\end{table}

All results where CPU time matters were obtained on a single thread of an AMD Ryzen 7 PRO 5850U using Julia 1.8.5 on Linux.
Differential equations are integrated with methods from the OrdinaryDiffEq.jl package \cite{DifferentialEquations-jl}.
To search fixed points of the differential equation, i.e. (local) minima of the loss, we use optimizers defined in Optim.jl \cite{Optim-jl} and NLopt.jl \cite{NLopt}.
Gradient descent (\code{Descent}) and Adam (\code{Adam}) are copied from Flux.jl \cite{Flux-jl} and defined directly in MLPGradientFlow.jl.

\begin{table}[h!]
    \footnotesize\centering
\begin{tabular}{l|ccc}
        name & needs Hessian & ode solver & works with minibatches\\\midrule
        Gradient Descent & no & yes (small step size) & yes \\
        Runge-Kutta 4 (RK4) & no & yes & no \\
        CVODE\_BDF & no & yes & no \\
        KenCarp58 & yes & yes & no \\
        Rodas5 & yes & yes & no \\\midrule
        Adam & no & no & yes \\
        (L)-BFGS & no & no & no  \\\
        SLSQP & no & no & no \\
        Newton(TrustRegion) & yes & no & no \\
    \end{tabular}
    \caption{\textbf{Method properties.} Differential equation solvers are listed on the upper half of the table, fixed point solvers on the lower half.
        There exist online (L)-BFGS methods (see e.g.\protect\cite{Moritz16,Bottou18}) that we did not consider here.
    }
\end{table}

\subsection*{Adaptive Ordinary Differential Equation Solvers are preferable over Gradient Descent or Adam}

\begin{figure}[!h]
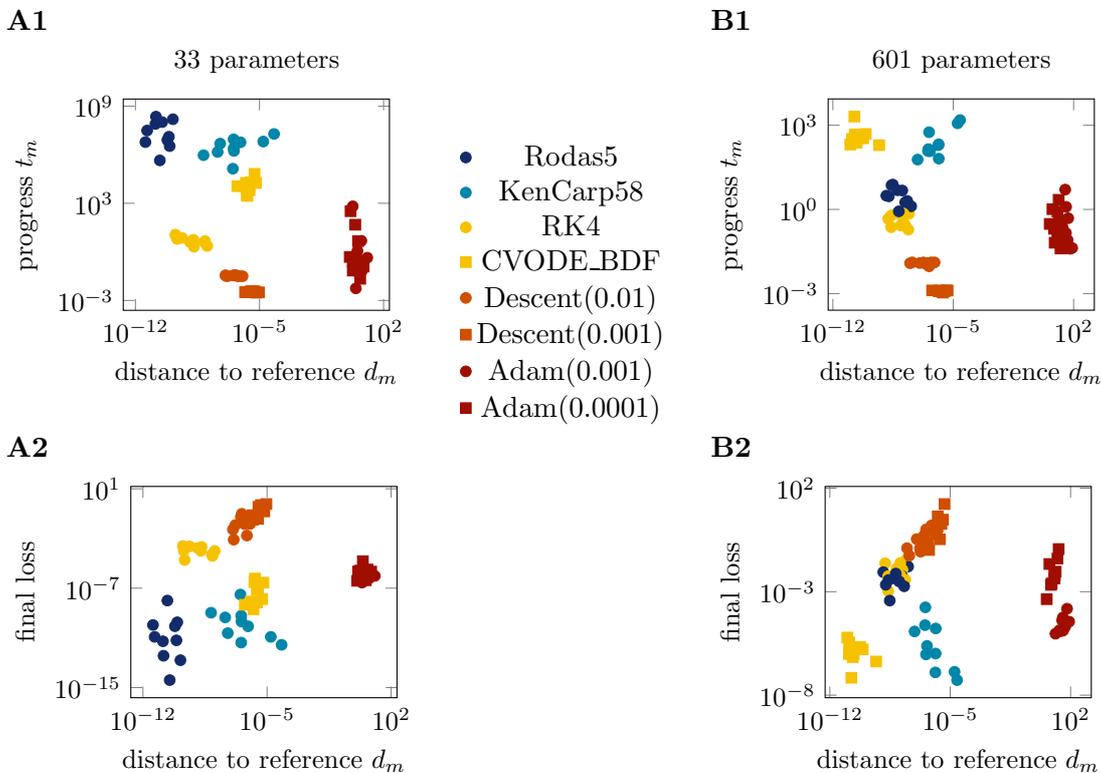

    \pgfplotsset{width = .33\textwidth}
    \centering
    \begin{tabular}{lll}
        \textbf{A1} & & \textbf{B1} \\
\end{tabular}
    \caption{\textbf{Efficiency and accuracy of different approximations to the gradient flow.}
    For given CPU time, higher-order methods Rodas5 and KenCarp58 reach high accuracy (low distance $d_m$ to gradient flow) and integrate the differential equation to a large $t_m$ (A1) and a low loss (A2) for networks with 33 parameters.
    For larger networks (601 parameters in B1 and B2), the gap between first order and higher order methods decreases because of the long durations for computing Hessians, but the differential equation solver CVODE\_BDF still excels.
    We used $10^4$ samples in 2 input dimensions in these simulations.
}\label{fig:ode}
\end{figure}

To compare different integration methods, we solved the gradient flow differential equation \autoref{eq:gradientflow} with a reference algorithm at tolerance $10^{-6}$ and kept the solution as a reference trajectory $(\theta^*(0), \theta^*(t_1), \ldots, \theta^*(T^*))$ at $10^6$ logarithmically spaced time points $(t_0 = 0, \ldots, t_{10^4} = T^*)$.
The reference algorithm was Rodas5 for the network with 33 parameters and CVODE\_BDF for the network with 601 parameters.
Next, we ran different integration methods $m$ for a fixed duration on a single CPU thread and kept their trajectories $(\theta^m(0), \theta^m(t_1), \ldots, \theta^m(T^m))$ at $10^3$ logarithmically spaced time points.
For each method $m$, we computed the distance to the reference target trajectory\linebreak $d_m = \frac1{10^3}\sum_{i = 1}^{10^3}\min_j\|\theta^*(t_j) - \theta^m(t_i)\|_2^2$, the progress $t_m = \arg\min_{t}\|\theta^*(t) - \theta^m(T^m)\|^2_2$ and the final loss at $\theta^m(T^m)$.
This was repeated for two different network sizes, and multiple random seeds for the dataset and the initial condition.
We find for small networks that higher-order methods Rodas5 and KenCarp58 are usually preferable over CVODE\_BDF, Runge Kutta 4 (RK4), Euler integration (Gradient Descent) or Adam and for larger networks CVODE\_BDF works best (\autoref{fig:ode}).

\begin{figure}[ht!]
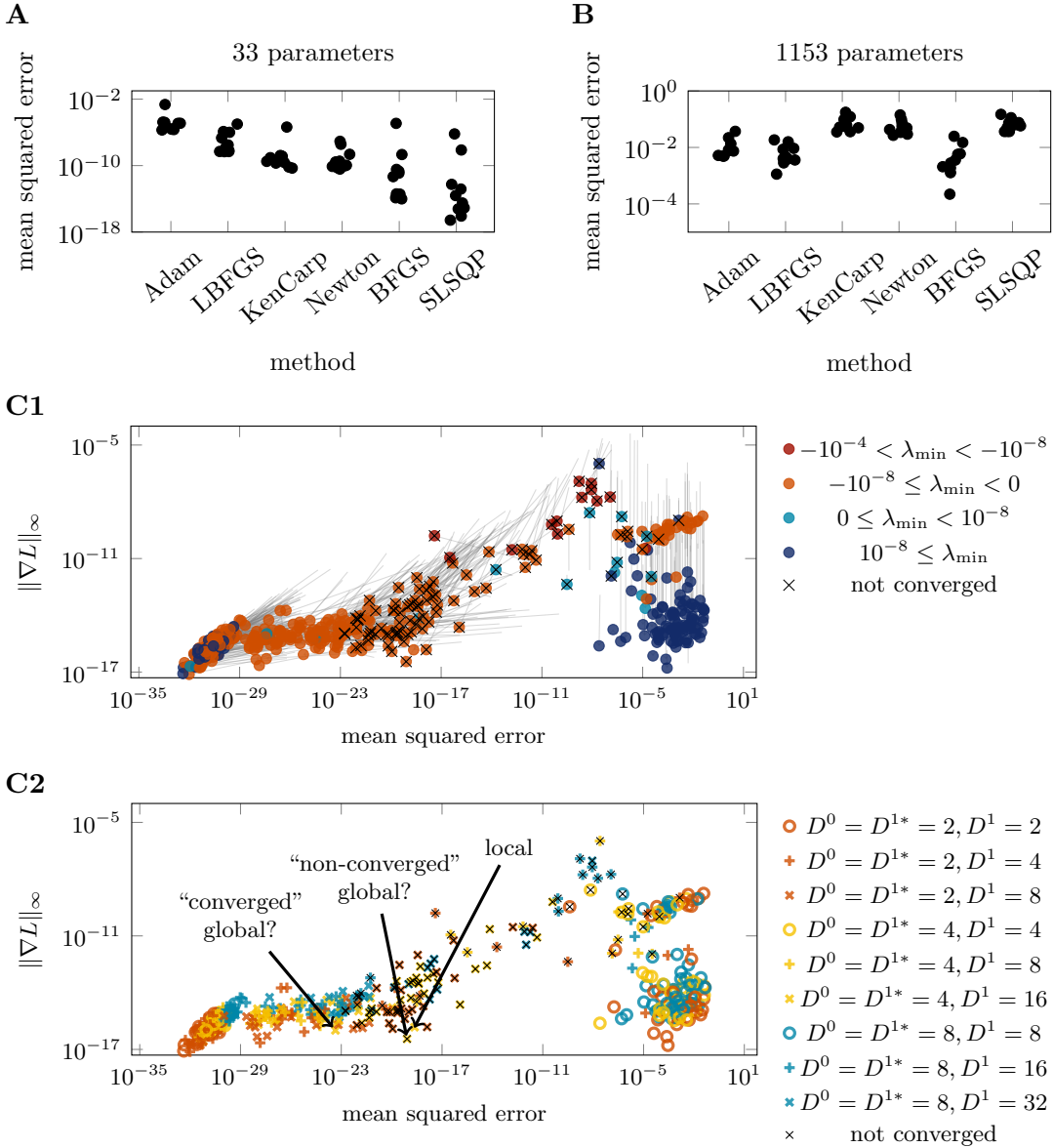

    \pgfplotsset{width = .42\textwidth, height = 3.5cm}
    \centering
\begin{tabular}{lll}
        \textbf{A} & & \textbf{B} \\
\end{tabular}
    \caption{\small\textbf{Finding fixed points efficiently.} For a fixed time budget, the approximate second order method SLSQP finds typically the lowest loss value for small networks (\textbf{A}), whereas BFGS is preferable for larger networks (\textbf{B}).
        In \textbf{C} we show simulation results, where we know that the smallest loss is at 0.
        The starts of the gray lines in \textbf{C1} show the gradient norm and loss value after $10^4$ steps of KenCarp58 and the points show the gradient norm, loss value and minimal eigenvalue of the Hessian (in color code) after at most 30 epochs of $10^4$ additional steps of Newton's method.
        We call a simulation ``probably converged'', if the loss found by Newton's method stayed the same or increased relative to the previous epoch.
        If this never happened within 30 epochs, we call the simulation ``not converged'' (marked with a black cross).
        In \textbf{C2} the show the same data with symbol- and color-code indicating the number of input dimensions $D^0$ the number of hidden neurons of the teacher $D^{1*}$ and the number of hidden neurons of the student $D^1$.
        Based on inspection of the parameters, we belief that the points marked ``global?'' will converge to a global minimum, if given more compute time, whereas the nearby point marked with ``local'' is probably stuck in a local minimum.
        Therefore, it is unclear, if the points marked as ``probably converged'' (no black cross) at losses between $10^{-30}$ and $10^{-16}$ reached a local minimum or if they did not fully converge to a global minimum, despite our stringent definition of convergence.
    }\label{fig:fixed_point}
\end{figure}

\subsection*{Finding Fixed Points}
Finding fixed points is surprisingly difficult even for small networks and datasets.
If we are only interested in finding low loss values within a given time budget, (approximate) second order optimizers like (L)BFGS, SLSQP or Newton's method are usually better than differential equation solvers, presumably because they do not spend time following accurately the trajectory of the gradient flow (\autoref{fig:fixed_point}A\&B).
To find fixed points efficiently, we therefore suggest switching to (approximate) second-order optimization after using differential equation solvers for a given time or iteration limit.

We investigated convergence in a teacher-student setup, where two-layer teachers ($L = 2$) are generated with independently and normally distributed parameters and the students have at least as many neurons in layer 1 as the teachers.
In this setting, the students can theoretically reach zero loss.
However, because we use double precision (64-bits) for all floating point numbers, the round-off errors for the difference between target values $y_i$ and predictions of the students $f_\theta(x_i)$ are on the order of $10^{-16}$.
Hence, the mean squared error is on the order of $(10^{-16})^2 = 10^{-32}$ for simulations that converge to a global minimum.

We observe, first, that it is indeed possible to reach loss levels on the order of $10^{-32}$ with KenCarp58, Newton's method and SLSQP.
BFGS usually stalls at loss levels around $10^{-14}$, unless the gradients are scaled with $\eta \gg 1$.
However, even with rescaling BFGS usually does not find global minima as accurately as Newton's method or SLSQP.
Second, we find that it can take very long to converge, even with efficient higher-order methods; almost 20\% of the simulations reported in \autoref{fig:fixed_point} did not fully converge within $10^4$ steps of KenCarp58 and $30\times 10^4$ steps of Newton's method.
Third, it is difficult to find a reliable convergence criterion.
Tiny gradient norms and smallest Hessian eigenvalues near zero do not imply that a fixed point is reached (see ``non-converged'' point that is probably on the path to a global minimum in \autoref{fig:fixed_point}C2).
And even if we call a point ``probably converged'', because no decrease of the loss is found in subsequent epochs of $10^4$ steps of Newton optimization, simulations may not have fully converged to a fixed point (see ``converged'' point that is probably on the path to a global minimum in \autoref{fig:fixed_point}C2).

\clearpage
\subsection*{Efficiency}
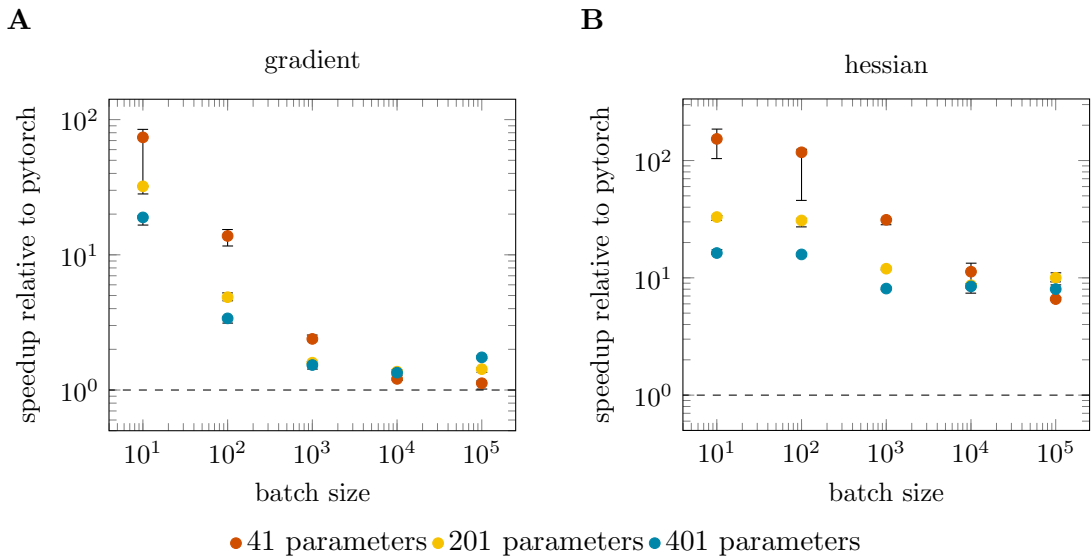
\begin{figure}[!h]
    \pgfplotsset{width = .45\textwidth}
    \centering
    \begin{tabular}{lll}
        \textbf{A} & & \textbf{B} \\
        \begin{tikzpicture}
\begin{axis}[xmode={log}, ymode={log}, ymin={0.5}, font={\small}, ylabel={speedup relative to pytorch}, legend columns={3}, legend style={draw={none}}, legend to label={pytorch_comparison}, xlabel={batch size}, title={gradient}]
    \addplot[scatter, only marks, error bars/y dir = both, error bars/y explicit, scatter src={explicit symbolic}, scatter/classes={41={mark={*}, color={rgb,1:red,0.8157;green,0.3059;blue,0.0}}, 201={mark={*}, color={rgb,1:red,0.9647;green,0.7608;blue,0.0}}, 401={mark={*}, color={rgb,1:red,0.0;green,0.5255;blue,0.6588}}}]
        table[row sep={\\}, y error plus={gyp}, y error minus={gyn}, x={N}, y={gmed}, meta={Nparams}]
        {
            N  Nparams  gmed  gmin  gmax  hmed  hmin  hmax  gyp  gyn  hyp  hyn  \\
            10  41
              73.91125769201926  31.985882431937462  84.68501231907128  152.87716859525483  103.9112534026493  185.37426730672294  10.773754627052014  41.9253752600818  32.49709871146811  48.965915192605536  \\
            100  41
              13.786967331523217  11.621852721296165  15.38503265352443  117.5517423463385  45.714283911795675  124.72591468059237  1.5980653220012133  2.165114610227052  7.174172334253868  71.83745843454282  \\
            1000  41
              2.3880028493172922  2.314319678893426  2.55123203675696  31.216141931033768  28.358031765010537  31.89355542019745  0.16322918743966763  0.07368317042386607  0.6774134891636834  2.8581101660232306  \\
            10000  41
              1.2085968704743837  1.1804898894648936  1.2949463312869898  11.278669305431986  8.92412397854421  13.329873197164048  0.08634946081260608  0.02810698100949005  2.0512038917320616  2.3545453268877754  \\
            100000  41
              1.1226716377345656  1.0165741399510895  1.166942661136488  6.590322614914293  6.416083806176112  6.801489759467716  0.04427102340192235  0.1060974977834761  0.21116714455342311  0.1742388087381812  \\
            10  201
              32.108059262619015  28.197941614976504  32.53277557396795  32.97780128566828  30.937428863007007  33.98964931029357  0.42471631134893784  3.9101176476425117  1.0118480246252943  2.040372422661271  \\
            100  201
              4.859734054294368  4.572734116733121  5.2263413717158524  30.82971922217552  27.187048337675108  32.126229754153414  0.3666073174214848  0.2869999375612471  1.2965105319778942  3.6426708845004114  \\
            1000  201
              1.5930332197720731  1.4285313979081684  1.6250680004652633  11.960128476149475  11.676068118422776  12.290682120367627  0.03203478069319021  0.16450182186390472  0.3305536442181527  0.28406035772669824  \\
            10000  201
              1.3672301889624014  1.285940485519908  1.41457173182536  8.648753050508457  8.176131744572297  8.813098007320503  0.04734154286295844  0.08128970344249353  0.1643449568120463  0.47262130593616014  \\
            100000  201
              1.427053816126882  1.3503082467563619  1.4565367847849102  10.01487025818506  9.242561100478405  11.03180064311422  0.02948296865802824  0.07674556937052013  1.0169303849291609  0.7723091577066548  \\
            10  401
              18.91227323746831  16.59705513740864  19.41861287499828  16.272657394892914  15.84463309782465  17.356123388044576  0.5063396375299725  2.3152181000596705  1.0834659931516626  0.42802429706826395  \\
            100  401
              3.3875544267801887  3.125583819496619  3.471558063094258  15.834271405310481  15.2863006015596  16.29565477962871  0.08400363631406949  0.2619706072835699  0.4613833743182276  0.5479708037508821  \\
            1000  401
              1.5308147477158818  1.486056406765233  1.5665715100653859  8.09834148128613  7.90706344983275  8.326394503260277  0.03575676234950409  0.04475834095064868  0.22805302197414612  0.19127803145338085  \\
            10000  401
              1.344114728095604  1.1794360625520584  1.3856466580609246  8.452009708854215  7.402404388257744  8.893933032110356  0.04153192996532051  0.1646786655435457  0.44192332325614103  1.0496053205964708  \\
            100000  401
              1.7445053566353699  1.703654237734113  1.788331417698228  8.015072806745616  6.575660372012492  8.705569637550061  0.043826061062858024  0.04085111890125681  0.6904968308044452  1.439412434733124  \\
        }
        ;
    \draw[dashed, black] ({rel axis cs:1,0}|-{axis cs:0,1}) -- ({rel axis cs:0,0}|-{axis cs:0,1});
    \legend{{41 parameters},{201 parameters},{401 parameters}}
\end{axis}
\end{tikzpicture}
  & &
        \begin{tikzpicture}
\begin{axis}[xmode={log}, ymode={log}, ymin={0.5}, font={\small}, ylabel={speedup relative to pytorch}, legend columns={3}, legend style={draw={none}}, legend to label={pytorch_comparison}, xlabel={batch size}, title={hessian}]
    \addplot[scatter, only marks, error bars/y dir = both, error bars/y explicit, scatter src={explicit symbolic}, scatter/classes={41={mark={*}, color={rgb,1:red,0.8157;green,0.3059;blue,0.0}}, 201={mark={*}, color={rgb,1:red,0.9647;green,0.7608;blue,0.0}}, 401={mark={*}, color={rgb,1:red,0.0;green,0.5255;blue,0.6588}}}]
        table[row sep={\\}, y error plus={hyp}, y error minus={hyn}, x={N}, y={hmed}, meta={Nparams}]
        {
            N  Nparams  gmed  gmin  gmax  hmed  hmin  hmax  gyp  gyn  hyp  hyn  \\
            10  41
              73.91125769201926  31.985882431937462  84.68501231907128  152.87716859525483  103.9112534026493  185.37426730672294  10.773754627052014  41.9253752600818  32.49709871146811  48.965915192605536  \\
            100  41
              13.786967331523217  11.621852721296165  15.38503265352443  117.5517423463385  45.714283911795675  124.72591468059237  1.5980653220012133  2.165114610227052  7.174172334253868  71.83745843454282  \\
            1000  41
              2.3880028493172922  2.314319678893426  2.55123203675696  31.216141931033768  28.358031765010537  31.89355542019745  0.16322918743966763  0.07368317042386607  0.6774134891636834  2.8581101660232306  \\
            10000  41
              1.2085968704743837  1.1804898894648936  1.2949463312869898  11.278669305431986  8.92412397854421  13.329873197164048  0.08634946081260608  0.02810698100949005  2.0512038917320616  2.3545453268877754  \\
            100000  41
              1.1226716377345656  1.0165741399510895  1.166942661136488  6.590322614914293  6.416083806176112  6.801489759467716  0.04427102340192235  0.1060974977834761  0.21116714455342311  0.1742388087381812  \\
            10  201
              32.108059262619015  28.197941614976504  32.53277557396795  32.97780128566828  30.937428863007007  33.98964931029357  0.42471631134893784  3.9101176476425117  1.0118480246252943  2.040372422661271  \\
            100  201
              4.859734054294368  4.572734116733121  5.2263413717158524  30.82971922217552  27.187048337675108  32.126229754153414  0.3666073174214848  0.2869999375612471  1.2965105319778942  3.6426708845004114  \\
            1000  201
              1.5930332197720731  1.4285313979081684  1.6250680004652633  11.960128476149475  11.676068118422776  12.290682120367627  0.03203478069319021  0.16450182186390472  0.3305536442181527  0.28406035772669824  \\
            10000  201
              1.3672301889624014  1.285940485519908  1.41457173182536  8.648753050508457  8.176131744572297  8.813098007320503  0.04734154286295844  0.08128970344249353  0.1643449568120463  0.47262130593616014  \\
            100000  201
              1.427053816126882  1.3503082467563619  1.4565367847849102  10.01487025818506  9.242561100478405  11.03180064311422  0.02948296865802824  0.07674556937052013  1.0169303849291609  0.7723091577066548  \\
            10  401
              18.91227323746831  16.59705513740864  19.41861287499828  16.272657394892914  15.84463309782465  17.356123388044576  0.5063396375299725  2.3152181000596705  1.0834659931516626  0.42802429706826395  \\
            100  401
              3.3875544267801887  3.125583819496619  3.471558063094258  15.834271405310481  15.2863006015596  16.29565477962871  0.08400363631406949  0.2619706072835699  0.4613833743182276  0.5479708037508821  \\
            1000  401
              1.5308147477158818  1.486056406765233  1.5665715100653859  8.09834148128613  7.90706344983275  8.326394503260277  0.03575676234950409  0.04475834095064868  0.22805302197414612  0.19127803145338085  \\
            10000  401
              1.344114728095604  1.1794360625520584  1.3856466580609246  8.452009708854215  7.402404388257744  8.893933032110356  0.04153192996532051  0.1646786655435457  0.44192332325614103  1.0496053205964708  \\
            100000  401
              1.7445053566353699  1.703654237734113  1.788331417698228  8.015072806745616  6.575660372012492  8.705569637550061  0.043826061062858024  0.04085111890125681  0.6904968308044452  1.439412434733124  \\
        }
        ;
    \draw[dashed, black] ({rel axis cs:1,0}|-{axis cs:0,1}) -- ({rel axis cs:0,0}|-{axis cs:0,1});
    \legend{{41 parameters},{201 parameters},{401 parameters}}
\end{axis}
\end{tikzpicture}
  \\
        \multicolumn{3}{c}{\ref{pytorch_comparison}}\\

    \end{tabular}
    \caption{\textbf{Speedup relative to pytorch.} The speedup factor (duration MLPGradientFlow/duration pytorch) for different network sizes is measured on a single CPU thread with multiple repetitions of gradient and Hessian computations using the Python module ``timeit'' and calling MLPGradientFlow from Python.}\label{fig:pytorch}
\end{figure}

We use a custom implementation of gradient and Hessian computations, taking inspiration from Bishop (1992) \cite{Bishop92} and SimpleChains.jl (\url{https://github.com/PumasAI/SimpleChains.jl}).
Relative to pytorch we observe a significant speedup, in particular for the computation of Hessians (\autoref{fig:pytorch}).

\subsection*{Finite versus Infinite Data}
\begin{figure}[h!]
    \centering
    \pgfplotsset{width = .75\textwidth, height = 6cm}
    \begin{tikzpicture}
\begin{axis}[xmode={log}, ymode={log}, ylabel={distance to infinite data solution}, xlabel={number of samples $N$}]
    \addplot[only marks]
        table[row sep={\\}, y={distance}, x={Nsamples}]
        {
            seed  Nsamples  duration  distance  rel_duration  \\
            1.0  1000.0  0.11482000350952148  0.13605742609955107  0.11066976591341796  \\
            2.0  1000.0  0.09379196166992188  0.22914808128787917  0.09040179520382756  \\
            3.0  1000.0  0.09617209434509277  0.48502479975651563  0.09269589656206517  \\
            4.0  1000.0  0.09305286407470703  0.21656112072546116  0.08968941273257276  \\
            5.0  1000.0  0.10911417007446289  1.0347245918921713  0.10517017323533095  \\
            6.0  1000.0  0.09225082397460938  26.634307706832114  0.08891636285086273  \\
            7.0  1000.0  0.10612607002258301  0.17384551973523396  0.10229007984428698  \\
            8.0  1000.0  0.08666205406188965  0.27389437671967626  0.0835296023641906  \\
            9.0  1000.0  0.0950009822845459  0.4342410945273892  0.09156711504632209  \\
            10.0  1000.0  0.08876895904541016  0.2491808943879612  0.08556035200886111  \\
            11.0  1000.0  0.10033202171325684  0.6335664818340663  0.09670546087124376  \\
            12.0  1000.0  0.10781192779541016  0.023998355618767127  0.10391500128113944  \\
            13.0  1000.0  0.09137701988220215  0.17817676999831172  0.08807414292920182  \\
            14.0  1000.0  0.09528422355651855  0.07368343707926811  0.09184011839337071  \\
            15.0  1000.0  0.10439705848693848  1.552787643049396  0.10062356446314512  \\
            16.0  1000.0  0.10464191436767578  0.11065777803092412  0.10085956988184792  \\
            17.0  1000.0  0.09117507934570312  0.10322798141817176  0.08787950165399123  \\
            18.0  1000.0  0.09557700157165527  2.0349554799915106  0.09212231377230648  \\
            19.0  1000.0  0.09584498405456543  2.0679891330760816  0.09238060986833563  \\
            20.0  1000.0  0.09946894645690918  0.30714860578477826  0.09587358198545591  \\
            1.0  10000.0  0.8393261432647705  0.006146170849274804  0.8089881985800609  \\
            2.0  10000.0  0.8693139553070068  0.059313367466075274  0.8378920832476368  \\
            3.0  10000.0  0.9567480087280273  0.05705438551767197  0.9221657805930928  \\
            4.0  10000.0  0.848952054977417  0.06313081546271751  0.8182661759653644  \\
            5.0  10000.0  0.9287958145141602  0.012417619852461249  0.895223935131831  \\
            6.0  10000.0  0.9339849948883057  0.13788437795626088  0.9002255494824312  \\
            7.0  10000.0  0.9333949089050293  0.08571972250365163  0.8996567925094132  \\
            8.0  10000.0  0.9069020748138428  0.030945396831905808  0.8741215577276837  \\
            9.0  10000.0  0.8493180274963379  0.04662761221054943  0.8186189201890341  \\
            10.0  10000.0  0.8724279403686523  0.17481599068847803  0.8408935114595912  \\
            11.0  10000.0  0.9072701930999756  0.0450249690849147  0.8744763701585281  \\
            12.0  10000.0  0.8791580200195312  0.01219751915775569  0.8473803283623591  \\
            13.0  10000.0  0.9295539855957031  0.02041722499160704  0.8959547016668601  \\
            14.0  10000.0  0.8199031352996826  0.013394059299372886  0.7902672468370793  \\
            15.0  10000.0  0.9583230018615723  0.17707644730477312  0.9236838446592571  \\
            16.0  10000.0  1.1030831336975098  0.053452586727110334  1.0632115350808151  \\
            17.0  10000.0  0.8936681747436523  0.015475984864031699  0.8613660048786709  \\
            18.0  10000.0  0.941072940826416  0.08322291262712819  0.9070572973817647  \\
            19.0  10000.0  0.8618209362030029  0.024047151057670097  0.8306699037938963  \\
            20.0  10000.0  0.9206929206848145  0.035859959284134255  0.8874139252389066  \\
            1.0  100000.0  9.937060117721558  0.0015377636765093266  9.577879145462756  \\
            2.0  100000.0  9.511639833450317  0.006929060821431528  9.167835931422847  \\
            3.0  100000.0  10.166351079940796  0.0032853174548574493  9.79888224892252  \\
            4.0  100000.0  10.241843938827515  0.005417042523658284  9.87164637334127  \\
            5.0  100000.0  9.797797918319702  0.002013997976229811  9.443650661424144  \\
            6.0  100000.0  10.366615056991577  0.003284793524681027  9.991907564927343  \\
            7.0  100000.0  10.844027042388916  0.007148082237566314  10.452063209007273  \\
            8.0  100000.0  9.890913963317871  0.0016143794098013955  9.53340097136797  \\
            9.0  100000.0  9.90416693687439  0.006724658728284567  9.546174908280756  \\
            10.0  100000.0  10.30655288696289  0.00900314260948005  9.934016377902815  \\
            11.0  100000.0  10.216934204101562  0.01735030059117108  9.847637015852808  \\
            12.0  100000.0  10.084747076034546  0.0004199784188144483  9.720227870470483  \\
            13.0  100000.0  9.83647108078003  0.006299187546851589  9.480925959332152  \\
            14.0  100000.0  10.460123062133789  0.002854540412925385  10.08203566738173  \\
            15.0  100000.0  10.000439882278442  0.008501275409943297  9.63896801058003  \\
            16.0  100000.0  9.797411918640137  0.028968402758917367  9.443278613933511  \\
            17.0  100000.0  10.75830888748169  0.006073597922311936  10.369443387999112  \\
            18.0  100000.0  10.412994146347046  0.001966324497527886  10.036610254400973  \\
            19.0  100000.0  10.112174987792969  0.0019133867086455415  9.746664383978748  \\
            20.0  100000.0  10.244343996047974  0.0013299048716471575  9.874056064500488  \\
        }
        ;
    \addplot[no marks, red]
        coordinates {
            (1000,0.3720762805790176)
            (10000,0.03852901961178712)
            (100000,0.003989733906002688)
        }
        ;
    \node[red] at (3e4, 4e-2) {$\propto 1/N$};
\end{axis}
\end{tikzpicture}
     \caption{\textbf{Infinite, standard Gaussian input data.} The accuracy of the finite-data solution improves with rate $1/N$.
    These results were obtained with 9 input dimensions.}\label{fig:infinite_data}
\end{figure}
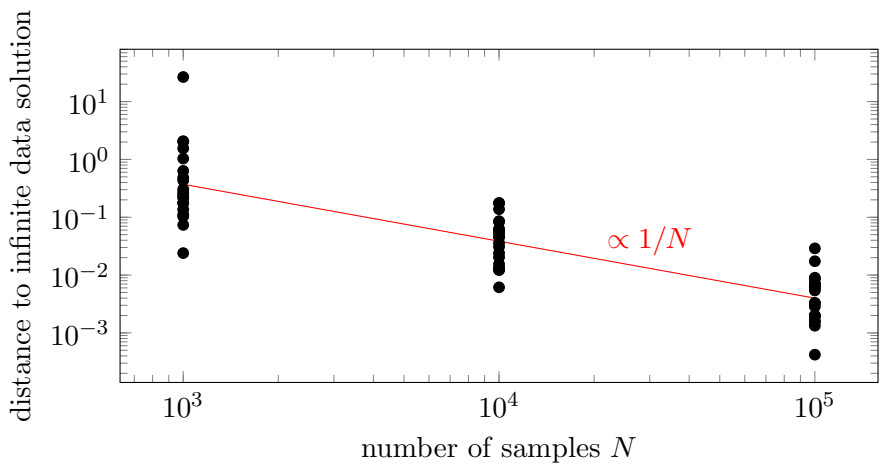
In a teacher-student setting with bias-free 2-layer networks and normally distributed input data, the loss function can be written in terms of one- to three-dimensional Gaussian integrals, independently of the input dimension \cite{Goldt19}.
For linear, relu and erf activation functions these integrals can be evaluated analytically, otherwise they can be computed numerically using cubature methods (we use Cuba.jl \url{https://github.com/giordano/Cuba.jl}).
However, accurately computing these integrals in every step of a differential equation solver is time consuming.
Therefore we fitted 3-layer perceptrons with 96 neurons in each hidden layer to datasets with approximately $10^5$ samples, such that they approximate the numerical integrals with high accuracy (typically, loss and gradients are within $10^{-5}$ Euclidean distance from the numerical integrals on test data, and Hessians are within $10^{-3}$ distance).
We recommend using these approximations for the differential equation solvers and switching to numerical integration, if fine-tuning of the fixed points is desired.
In contrast to existing differential equation solvers for this setting (e.g. \cite{Goldt19}), this package supports more activation functions and provides also Hessians, which enables the usage of higher-order solvers.

As an alternative to numerical integration or approximations thereof, one can sample datasets with normally distributed input.
But the accuracy of the finite-data solution improves only with a rate inversely proportional to the size of the dataset (\autoref{fig:infinite_data}), whereas the computational cost increases linearly with the size of the dataset.
In particular in high input dimensions, the numerical integrals and its approximations are preferable over approximations with finite data sets.

\section*{Conclusions}
We described the software package MLPGradientFlow.jl, a tool to integrate the gradient flow differential equation of the loss function of neural networks.
We illustrated the advantages and drawbacks of different methods with toy examples.
The package is mainly targeted at small networks with a few thousand parameters, at most.
For larger networks, the time and space requirements of computing and inverting Hessians becomes impractical and Hessian-free methods should be considered (e.g. \cite{Pearlmutter94} or integrators based on Krylov methods \cite{DifferentialEquations-jl}).
The largest setting we considered so far is a network with 79'510 parameters and 70'000 samples in 784 input dimensions, which we integrated with CVODE\_BDF (see ``scripts/mnist.jl'' in \url{https://github.com/jbrea/MLPGradientFlow.jl}).

\bibliographystyle{unsrt}

\end{document}